\pdfoutput=1

\documentclass{article}

\usepackage[preprint]{neurips_2026}

\usepackage[utf8]{inputenc}
\usepackage[T1]{fontenc}
\usepackage{hyperref}
\usepackage{url}
\usepackage{booktabs}
\usepackage{amsmath}
\usepackage{amsfonts}
\usepackage{amssymb}
\usepackage{graphicx}
\usepackage{nicefrac}
\usepackage{microtype}
\usepackage{xcolor}
\usepackage{caption}  %

\bibpunct{[}{]}{;}{a}{,}{,}

\newcommand{\secref}[1]{Sec.~\ref{#1}}

\title{Decomposing LLM-Judge Uncertainty to Target Expert Labels}

\author{%
  Ryan Lail \\
  Composo AI \\
  \texttt{ryan@composo.ai} \\
}

\begin{document}

\maketitle

\begin{abstract}
  An LLM judge evaluates outputs at scale. Experts should label only where it is
  least sure. Its natural escalation signal conflates two uncertainties:
  \emph{aleatoric}, real disagreement in the expert pool, which labels cannot
  reduce, and \emph{epistemic}, the judge's ignorance, which labels do reduce. A
  small Bayesian model separates them: a regression on labels already collected
  learns how far to trust a black-box judge's prediction. Both components follow as
  simple formulas, with no sampling or further judge calls. The components isolate on a
  real LLM judge against exactly known truth, and stated confidence is no guide to
  its actual error. On real human disagreement (ChaosNLI) the epistemic ranking removes
  83\% more error than total uncertainty for the same expert labels, though simply
  escalating the least-labelled items does as well there. We demonstrate we can estimate where a judge is ignorant rather than where
  experts genuinely disagree, and propose using this to direct expert labelling.
  Code and data are available at
  \url{https://github.com/composo-ai/judge-uncertainty-decomposition}.
\end{abstract}

\section{Introduction}
\label{sec:intro}

LLMs increasingly serve as judges: evaluators substituting for expert labels at
scale. Judges are stochastic. Averaging repeated calls is the
standard remedy \citep{lail2026}, but treats all variation as noise, when only
part of it is. The \emph{pool} of human experts the judge stands in for may read an
item differently, so the judge should predict a \emph{distribution} over labels, not
a single verdict \citep{baan2022,leonardelli2025}. Its spread is \emph{aleatoric}
uncertainty, a property of the item no further labelling removes. The judge may also
be ignorant of where the pool's distribution sits: \emph{epistemic} uncertainty.

This paper asks whether the two can be disentangled for a black-box judge, from
its outputs and the labels already collected.

\paragraph{Contributions.}
(1) \emph{A model of judge uncertainty separating the two components}: a small
regression on already-collected labels learns how far to trust the judge's
prediction on each item. Both components follow as simple formulas with no further
judge calls. The entropy of the predicted distribution, the
natural escalation score, is exactly the two components summed, which is why ranking
on it conflates ignorance with disagreement.
(2) \emph{A controlled test against exact ground truth}: a frontier LLM judges dice
whose true bias and pool disagreement are known, with the judge black-box.
The components isolate, and the judge's stated confidence, taken at face value,
is no guide to its actual error.
(3) \emph{A real-world check on human disagreement}: on ChaosNLI the epistemic
ranking recovers more label-budget value than total uncertainty, though not more
than a count-based rule.

\section{Related work}
\label{sec:related}

Calibration to majority labels is ill-posed under genuine disagreement
\citep{baan2022,plank2022}. LeWiDi-2025 \citep{leonardelli2025} standardises evaluation
against the pool's distribution. Distributional judging \citep{chen2025,wang2025}
supplies the prediction our model consumes, without per-item uncertainty about it.
Systems that escalate on undivided judge-side uncertainty target a single verdict
\citep{jung2025,sheng2025,badshah2026}. \citet{bachar2026} carry the distinction only as
self-reported flags, and \citet{dorner2025} prove an aggregate limit on judge value. Our
framing and the classical decomposition are the same object, not rivals
\citep{kotelevskii2025}, and our epistemic score is BALD's
\citep{houlsby2011,bickfordsmith2023,nguyen2019}. \citet{dubois2026} fit generalised
linear models to grader scores: theirs separates grader severity offsets from residual
disagreement, while ours separates irreducible pool disagreement from our own
ignorance of where the pool sits. The mechanism is theirs. The transfer, failure characterisation,
and judge-side measurement are ours. None asks which labels to buy.

\section{Model}
\label{sec:model}

The model builds up in three levels: a single annotator (level 0), the pool of
annotators (level 1), and our knowledge of the pool (level 2).

\paragraph{Levels 0 and 1: an annotator, and the pool.}
Annotator $a$ in a pool with weights $w_a$ holds a label distribution
$\pi_a(x)$ over the $K$ classes, and a label is a draw from it. A label from a
randomly drawn annotator is then a draw from the pool's \emph{population label
distribution} $p_x = \sum_a w_a \pi_a(x)$. Its entropy $H(p_x)$ (or Gini
$G(p_x) = 1 - \lVert p_x \rVert^2$) is the \emph{aleatoric} component: a
property of the pool, unreduced by labels. We model the pool at this
marginal level (real-valued scores: App.~\ref{app:model}).

\paragraph{Level 2: our knowledge of the pool.}
$p_x$ is never observed, so we hold a belief about it. An estimate's realised error is a
divergence $D_x = d(\hat p_x, p_x)$. \emph{Epistemic} uncertainty is a
prediction of $D_x$ made without observing $p_x$. $d$ is the Manhattan distance
$\sum_j |\hat p_j - p_j|$, running 0 (exact) to 2 (probability on disjoint
classes), standard for categorical soft labels \citep{leonardelli2025}.
Correctly predicting a genuinely split pool gives $D_x = 0$: disagreement is not
error.

\paragraph{The estimator.}
The level-2 belief is a Dirichlet centred on an LLM judge's predicted distribution
$\hat q_x$, weighted by a \emph{trust} $\lambda_x$ in pseudo-label units. The
$n_x$ labels in hand, tallied $c_x$, update it in closed form to
$\mathrm{Dir}(\alpha_x)$ with mean
$m_x = \alpha_x / \alpha_{0,x}$ and \emph{total evidence}
$\alpha_{0,x} = \lambda_x + n_x$, trust plus labels. Trust is fitted log-linearly
on item features, among them embedding distance/density to labelled items, the
\emph{familiarity channel}, a signal not from the judge. Ordinary labels are the
only training signal (full displays and likelihood: App.~\ref{app:model}). The
judge's prediction is never fitted: the judge does the perceptual work, our model
decides how far to believe it.

\paragraph{The decomposition.}
A fresh label's uncertainty splits as
\begin{equation*}
  H(m_x) = \mathbb{E}[H(p_x) \mid c_x] + I(y; p_x \mid c_x),
\end{equation*}
with $y$ one further label. The aleatoric term is closed-form
(App.~\ref{app:model}). The epistemic term $I_x$ is BALD's acquisition score
\citep{houlsby2011}, and total uncertainty $H(m_x)$ is the escalation baseline.
Entropy sees only the shape of $m_x$, not the evidence behind it: with equal
total evidence the two rankings nearly agree, and where evidence varies they
come apart, up to full inversion (App.~\ref{app:model}).

\section{Experimental setup}
\label{sec:setup}

Two experiments share one real LLM judge, Azure OpenAI's \texttt{gpt-5.6-terra},
queried 18 August 2026.

\paragraph{Experiment 1: dice (exactly known truth).}
500 three-sided dice with biases $p_i$ carry descriptions in four tiers (none,
weak, directional, strong), assigned independently of bias. From the description
alone the judge predicts the 100-roll distribution (three prompt variants,
0--100 stated confidence), with 0--10 revealed rolls as labels. With $p_i$ known,
true disagreement is its entropy, true error the posterior's Manhattan
distance from it. The regression is given each die's tier as its familiarity
feature. We measure whether each estimated component tracks its ground truth, and
whether stated confidence does.

\paragraph{Experiment 2: ChaosNLI (real human disagreement).}
3,113 items \citep{nie2020}, each a premise--hypothesis sentence pair labelled
entailment, neutral, or contradiction by 100 annotators. The judge predicts the
annotator distribution. Labels are half-split per item,
fitting and acquisition on one half, evaluation the other: evaluation truth is
never touched by training, and disjoint 50-label halves set a 0.158 noise ceiling
(Manhattan). We measure the estimation error each escalation rule removes for a fixed
escalation budget $B$, the fraction of items sent for expert labels: every rule
escalates its top-ranked $B\%$, so rules compete purely on which items they pick.

\paragraph{Escalation rules.}
\emph{Ours} ranks by the estimated epistemic component: mutual information $I_x$
under log loss and, under squared error, the one-label value $\Delta_x$, with
posterior spread $v_x$ the withhold variant (App.~\ref{app:model}). Experiment 2
reports all three. Baselines: \emph{entropy} (output or
posterior), \emph{verbalised confidence}, \emph{fewest-labels-first},
\emph{random}, \emph{oracle}.

\section{Results}
\label{sec:results}

\subsection{Simulations: the estimator on synthetic pools with known truth}
\label{sec:sims}

Two simulations precede the judge experiments: one calibrated, one
misspecified (setup and the misspecified results: App.~\ref{app:sims}). The
calibrated one draws worlds from the \secref{sec:model} model itself, so the
value of every escalation choice is exactly computable. When items differ only
in how many labels they already hold, ranking by total uncertainty forfeits a
third of the error reduction the same labels could have bought, while
epistemic ranking stays at oracle level. When evidence is equal the gap is
exactly zero, as the \secref{sec:model} claim requires. Posterior spread alone
captures nearly all of that gain, so the takeaway is simple: escalate on the
epistemic component (numbers: App.~\ref{app:sims}, Figure~\ref{fig:sims}).

\subsection{Experiment 1: a real LLM judge on dice with exactly known truth}
\label{sec:exp1}

\begin{table}[t]
\begin{minipage}[t]{0.40\linewidth}
    \caption{Component isolation: Spearman rank correlation of each estimate with
    the true quantities, 500 dice. The truths are themselves anti-correlated
    ($-0.23$), so the separation is not inherited from the world. Full grid:
    App.~\ref{app:exp1}.}
  \label{tab:iso}
  \centering
  \footnotesize
  \setlength{\tabcolsep}{3.5pt}
  \begin{tabular}{lcc}
    \toprule
    estimate & vs. disagreement & vs. error \\
    \midrule
    aleatoric & \textbf{0.45} & $-0.12$ \\
    epistemic ($I$) & 0.08 & \textbf{0.66} \\
    \bottomrule
  \end{tabular}
\end{minipage}\hfill
\begin{minipage}[t]{0.58\linewidth}
  \setlength{\tabcolsep}{3.5pt}
    \caption{Escalation on ChaosNLI at budget $B=10\%$ (the top tenth of items
    escalated), Manhattan units of error removed.
    All seven rules: App.~\ref{app:exp2}.}
  \label{tab:esc}
  \centering
  \footnotesize
  \begin{tabular}{lcc}
    \toprule
    rule & value & vs. entropy \\
    \midrule
    epistemic ($I$-ranking) & \textbf{15.0} & $+6.8$ $[+1.7, +11.9]$ \\
    fewest-labels-first & 14.3 & $+6.1$ $[-0.0, +12.4]$ \\
    entropy (total uncertainty) & 8.2 & --- \\
    \bottomrule
  \end{tabular}
\end{minipage}
\end{table}

\begin{figure}[t]
\noindent
\begin{minipage}[c]{0.55\linewidth}
  \centering
  \includegraphics[width=\linewidth]{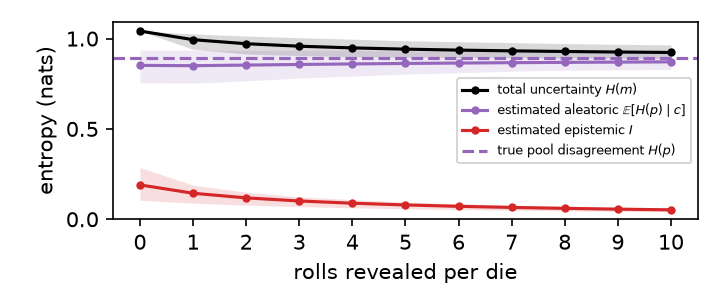}
  \caption{The decomposition in motion (500 dice, 200 draws, 2.5--97.5 bands).
    The aleatoric estimate holds at 0.85--0.87 against the true 0.892 (dashed),
    the epistemic falls 0.190 to 0.052, and total uncertainty falls to the
    aleatoric floor: labels drain ignorance, not disagreement.}
  \label{fig:convergence}
\end{minipage}\hfill
\begin{minipage}[c]{0.42\linewidth}
  \centering
  \footnotesize
  \setlength{\tabcolsep}{4pt}
  \captionof{table}{The confidence pathology by tier: stated confidence
    (0--100) is non-monotone in judge error. Full four-column table:
    App.~\ref{app:exp1}.}
  \label{tab:conf}
  \begin{tabular}{lcc}
    \toprule
    tier & judge error & stated confidence \\
    \midrule
    none        & 0.53  & \textbf{90.9} \\
    weak        & 0.54  & \textbf{0.4}  \\
    directional & 0.41  & 59.1           \\
    strong      & 0.007 & 14.8           \\
    \bottomrule
  \end{tabular}
\end{minipage}
\end{figure}

On a real LLM judge the two components are separable: each estimated component
tracks its own ground truth and not the other's (Table~\ref{tab:iso}).
Figure~\ref{fig:convergence} shows it in motion.

Stated confidence fails as an escalation signal: near-certainty
with no information, near-zero when merely told the dice are weighted, calibrated
only near complete information (Table~\ref{tab:conf}). At item level it
correlates \emph{positively} with error (Spearman $+0.22$), and escalating on low
confidence does no better than random (App.~\ref{app:exp1}).

Dice are the wrong world for measuring escalation gains, which is why the
decision-level claim waits for Experiment 2: this judge predicts uniform
exactly when ignorant, so ranking by total uncertainty nearly coincides with
ranking by ignorance, and between-rule differences drown in single-draw noise
(App.~\ref{app:exp1}). ChaosNLI supplies the regime where the two come apart.

\subsection{Experiment 2: escalation on real human disagreement (ChaosNLI)}
\label{sec:exp2}

On ChaosNLI the judge is confidently wrong again: its predicted annotator
distributions miss at triple the noise ceiling, the irreducible error from
annotator sampling (\secref{sec:setup}), while its stated confidence
averages above 90 on the 0--100 scale. Neither confidence nor prediction
entropy tracks error (App.~\ref{app:exp2}).

The regression finds near-uniform trust, each prediction worth about five
labels, unsurprising for short crowdsourced NLI pairs, which give the
familiarity features little to exploit. But label counts range 0--10 by design,
leaving total evidence $\alpha_0$ heterogeneous where $\lambda$ is flat (the
\secref{sec:model} condition).

On real human disagreement the epistemic ranking recovers more label-budget
value than total uncertainty, though a count-based rule does as well here.
For the same label budget, epistemic ranking removes 15.0 units of error
against entropy's 8.2 (Table~\ref{tab:esc}): 83\% more per expert label. The
gap widens at larger budgets and is absent at smaller ones. The rules split into
count-aware and count-blind clusters, and the count heuristic is not beaten
(App.~\ref{app:exp2}).

The \secref{sec:model} mechanism, measured in \secref{sec:sims}, is present on
real data. Separating our estimator from the count heuristic needs items genuinely
differing in judge familiarity, with dense labels to score. No public
dataset offers both: the field's missing dataset rather than this paper's missing
experiment.

\section{Limitations and conclusion}
\label{sec:limitations}

We demonstrate the full method only on synthetic dice, where the truth is
exactly known. On ChaosNLI the judge was about equally unfamiliar with every
item, so its fitted trust ($\lambda$) came out nearly flat, and only the
label-count side of the model was really tested on real data. That side
worked, removing nearly twice as much error as entropy ranking at the same
budget (\secref{sec:exp2}). But when trust is flat our ranking follows label
counts, exactly what the fewest-labels-first baseline does, so the baseline
ties us. The remaining limits are more ordinary. Each experiment used one
judge model and one prompt design. Simulated effect sizes depend on the
generator (\secref{sec:sims}). Our intervals resample labels over a fixed
item set and say nothing about new items. We have no defence against a
deceptive judge that games the familiarity signals behind its trust. A merely
worthless signal reduces the method to ranking by label counts, still
${\sim}32$pp better than entropy ranking in the simulations
(\secref{sec:sims}).

Escalate the items the judge is ignorant about, not the items annotators
genuinely disagree on. The two can be separated with the labels already
collected. No new data are needed. Escalating on ignorance removed more error
per expert label than escalating on total uncertainty, though not more than a
fewest-labels-first rule. Overall, this demonstrates a first step towards
decomposing epistemic and aleatoric uncertainty in LLM judges. The decisive
test is on data where judge familiarity
genuinely varies, with the comparison against fewest-labels-first stated before
any results are seen.

\label{sec:refs}
\bibliographystyle{plainnat}
\bibliography{refs}

\appendix
\section*{Appendices}

\section{Supporting simulations in full}
\label{app:sims}

Terms used for escalation results here and in the body: a rule's \emph{regret} at
budget $B$ is its top-$B\%$ selection's shortfall below the best possible, in
percentage points (pp). \emph{Advantage} is the difference of two rules' regrets.
Escalation \emph{value} is the reduction in error from purchased labels, in
Manhattan units, summed over the escalated set. Real-judge numbers are means over
200 label re-draws with 95\% intervals, paired within re-draw.

\begin{figure}[h]
  \centering
  \includegraphics[width=\linewidth]{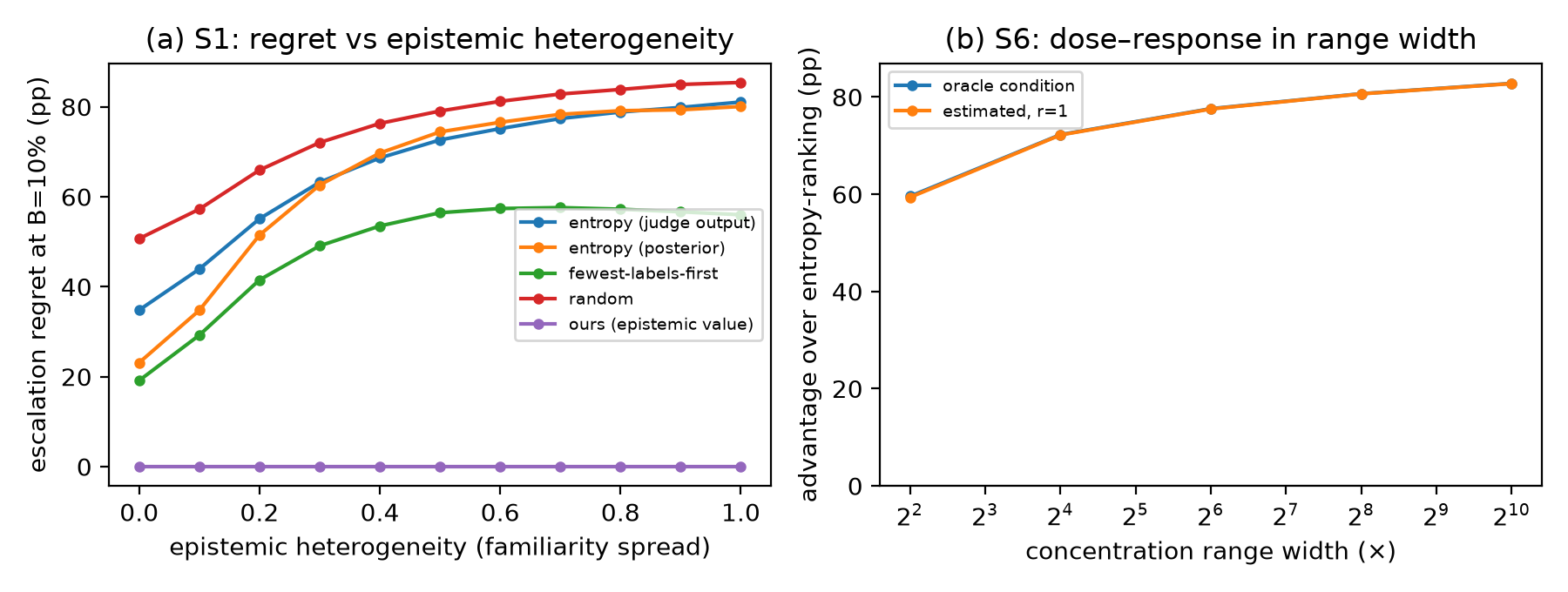}
  \caption{(a) Escalation regret at $B=10\%$ as the spread of item trust grows.
    Entropy-ranking (on the judge's output, or on the posterior with labels folded in)
    degrades from 35pp toward random's 85pp, fewest-labels-first captures part of the gap and
    plateaus, and ranking by estimated label value stays at oracle level. (b) The effect does not
    require an extreme world: as the trust range narrows from $\times 256$ to $\times 4$, the
    advantage over entropy-ranking declines only from 81pp to 60pp, and the estimated-trust
    condition matches the oracle condition at every width.}
  \label{fig:sims}
\end{figure}

\paragraph{Setup.}
In the \emph{calibrated} simulation (2,000 items), the world is drawn from the
\secref{sec:model} model with the judge's prior correct by construction, so the value of
every escalation decision is exactly computable. The spread of item trust, the spread of
pool disagreement, and the correlation between them are steerable axes, and each item
reveals 0--10 labels. In the \emph{misspecified} simulation, the world is drawn truth-first,
the judge's prediction quality varies with familiarity, and its stated confidence is
corrupted by a knob $\gamma$: honest at $\gamma=0$, information-free
at $\gamma=1$, inverted at $\gamma=1.5$. There the estimator's model is wrong by
construction and rules are scored against an oracle that knows the realised world, so only
between-rule comparisons are meaningful.

\paragraph{Calibrated simulation: all four findings.}
Pass/fail criteria: advantage $\ge 15$pp at the criterion cell and oracle tracking
Spearman $\ge 0.9$, measured 80.9pp (seed-level CI 80.6--81.2) and 1.0.
(1) The failure needs no exotic regime: with identical prior trust and only revealed counts
(0--10) differing, entropy-ranking loses 34.8pp. In the true null (identical trust
\emph{and} identical label count), the advantage measures 0.0pp, as the \secref{sec:model}
claim requires. (2) Entropy-ranking approaches random as trust spread grows (81pp regret vs
random's 85pp at the widest setting), because entropy is blind to evidence and, through
revealed counts, anti-correlated with label value: the claim's inversion, realised.
(3) There is no estimator-quality threshold, and adverse correlation does not close the gap:
with a worthless familiarity signal the regression still beats entropy-ranking by 32pp
(revealed counts alone carry the signal), rising monotonically to 81pp as the signal
improves, and when contentious items are also the unfamiliar ones (correlation 0.8, the
realistic direction), the advantage is still 50pp. (4) Ranking by posterior spread captures
${\sim}98\%$ of the gain ($v$-ranking within 1.2--1.8pp of $\Delta$-ranking everywhere). The
finer withhold-versus-purchase distinction (App.~\ref{app:model}) is second-order for
ranking.

\begin{figure}[t]
  \centering
  \includegraphics[width=\linewidth]{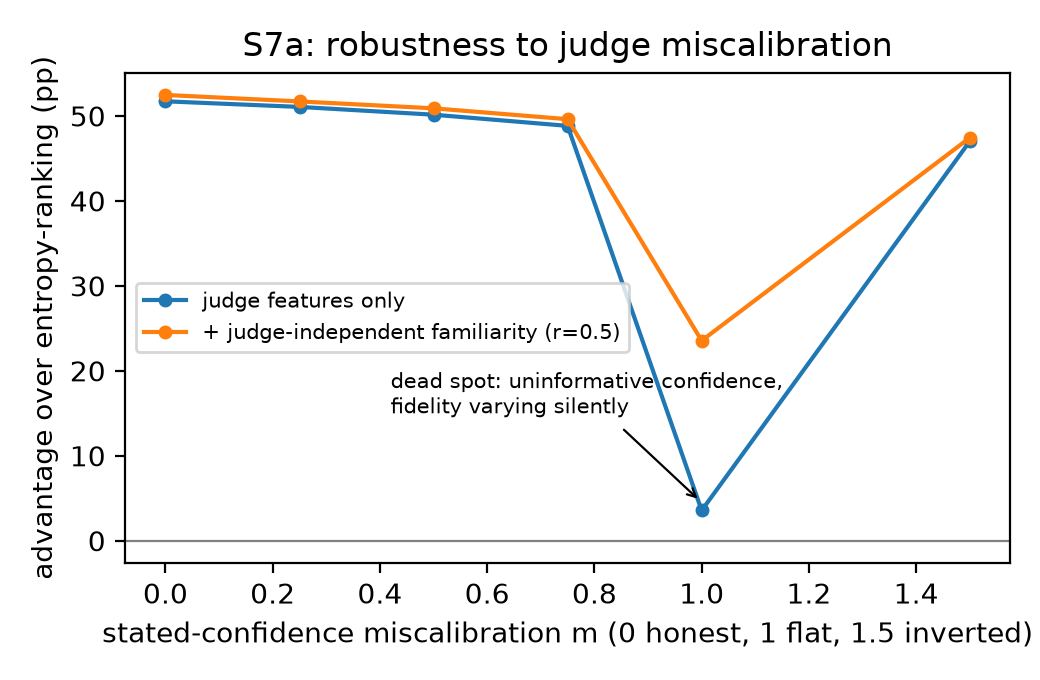}
  \caption{Advantage over entropy-ranking at $B=10\%$ in the misspecified simulation, as
    stated-confidence corruption $\gamma$ varies (the axis label reads $m$, this paper's
    earlier name for the knob). With judge-supplied features only (blue), the advantage is
    flat at ${\sim}50$pp from honest through badly miscalibrated ($\gamma \le 0.75$) and at
    fully inverted confidence ($\gamma=1.5$), where the regression re-learns the mapping and
    exploits the inversion (fitted coefficient on stated confidence flips from $-0.63$ to
    $+1.09$). The one dead spot is $\gamma=1$: confidence carrying no information while
    prediction quality varies silently (3.6pp). A judge-independent familiarity signal of
    middling quality (orange) recovers most of it (23.5pp) and costs nothing elsewhere.}
  \label{fig:robust}
\end{figure}

\paragraph{Misspecified simulation: the predictions as scored.}
The prediction that a regression using only revealed-label counts survives all tested
corruptions was refuted at $\gamma=1$ and confirmed elsewhere: the collapse requires both
uninformative confidence and silently varying quality. The prediction that the regression would
monotonically down-weight stated confidence as corruption grows was refuted in its
monotone form: the fitted coefficient grows through $\gamma=0.75$, reaches zero at
$\gamma=1$, and flips sign at inversion. The regression re-learns the mapping rather than
gradually distrusting it (Figure~\ref{fig:robust}). These magnitudes are
generator-dependent: they do not predict real-world effect sizes.

\section{Model details: the estimator, fitting likelihood, closed forms, and the real-valued formulation}
\label{app:model}

\paragraph{The estimator in full.}
An LLM judge supplies a prediction $\hat q_x$ of the pool. The belief, the labels
in hand ($n_x$ of them, class tally $c_x$), and the trust, fitted log-linearly on
item features, are
\begin{equation*}
  p_x \sim \mathrm{Dir}(\lambda_x \hat q_x), \qquad
  c_x \sim \mathrm{Multinomial}(n_x, p_x), \qquad
  \log \lambda_x = \beta_0 + \beta^\top z_x,
\end{equation*}
with $z_x$ collecting variant spread, stated confidence, the label count $n_x$
itself, and embedding distance/density to labelled items. The coefficients $\beta$
are the only fitted quantities, and the log keeps $\lambda_x$ positive. The posterior
is
\begin{equation*}
  p_x \mid c_x \sim \mathrm{Dir}(\alpha_x), \qquad
  \alpha_x = \lambda_x \hat q_x + c_x, \qquad
  \alpha_{0,x} = \lambda_x + n_x,
\end{equation*}
labels and trust in one currency: updating is addition.

\paragraph{The real-valued formulation.}
Annotator $a$ holds a latent score $\mu_a(x) \in [0,1]$, the score they would settle on
given unlimited care, and an observed label adds within-annotator noise, which we take to
be Gaussian,
\begin{equation*}
  y = \mu_a(x) + \varepsilon, \qquad \varepsilon \sim \mathcal{N}(0, \sigma_a^2),
\end{equation*}
where $\sigma_a^2$ is the variance of a fresh rating of the same item by the same annotator.
A pool of annotators $a = 1, \dots, M$ with weights $w_a$ induces a population score
distribution with mean and spread
\begin{equation*}
  m(x) = \sum_a w_a \mu_a(x), \qquad
  \tau^2(x) = \sum_a w_a (\mu_a(x) - m(x))^2,
\end{equation*}
where $\tau^2(x)$ is genuine inter-annotator disagreement, a property of the pool itself,
unreduced by collecting further labels.

Neither $\mu_a(x)$ nor $\tau^2(x)$ is directly observable: a label carries the annotator's
own noise with it. Averaging labels still recovers $m(x)$, since that noise is centred, but a
label's variance is $\tau^2(x) + \bar\sigma^2$ with $\bar\sigma^2 = \sum_a w_a \sigma_a^2$, so
one label per (item, annotator) pair cannot separate genuine disagreement from
within-annotator noise. It need not: both are aleatoric, and every quantity in the paper uses
only their sum.

For real-valued scores, the split displayed in \secref{sec:model} is the three-way variance
decomposition
\begin{equation*}
  \mathrm{Var}(y \mid x, \mathcal{D}) =
  \underbrace{\bar\sigma^2}_{\text{within-annotator}} +
  \underbrace{\mathbb{E}[\tau^2 \mid \mathcal{D}]}_{\text{between-annotator}} +
  \underbrace{\mathrm{Var}(m \mid \mathcal{D})}_{\text{epistemic}},
\end{equation*}
whose first two terms are aleatoric (irreducible for a single fresh label) and whose third
alone shrinks with labels. The entropy display of \secref{sec:model} is its categorical
counterpart, with the first two terms merged by the Level-0 identifiability constraint above.

\paragraph{The fitting likelihood.}
The maximum-likelihood objective for $\beta$ is the Dirichlet--multinomial likelihood
\begin{equation*}
  \mathcal{L}(\beta) = \prod_{x: n_x \ge 1}
  \mathrm{DirMult}(c_x \mid n_x, \lambda_x(\beta) \hat q_x),
\end{equation*}
a generalised linear model with log link on $\lambda$ and Dirichlet--multinomial response
(beta--binomial regression generalised past two classes). A handful of
coefficients against a few hundred labelled items is well-posed: only the trust
scale is learned.

\paragraph{Closed form of the aleatoric term.}
$\mathbb{E}[H(p_x) \mid c_x] = \psi(\alpha_{0,x}+1) - \sum_j m_{x,j} \psi(\alpha_{x,j}+1)$,
with $\psi$ the digamma function.

\paragraph{The derived scores.}
Under squared error the decomposition yields two decision quantities: the posterior spread
$v_x = G(m_x)/(\alpha_{0,x}+1)$, the expected squared error of the current estimate (the
\emph{withhold} quantity), and the one-label value
\begin{equation*}
  \Delta_x = \frac{G(m_x)}{(\alpha_{0,x}+1)^2},
\end{equation*}
the expected reduction in that error from one further expert label (the \emph{purchase}
quantity). Withholding and purchasing optimise different scores \citep{bickfordsmith2023},
both distinct from entropy. In this model class the two rank items nearly identically
(\secref{sec:sims}), and the claim the paper tests is the two-way split, epistemic
against total.

\paragraph{When entropy-ranking is safe.}
Entropy sees only the shape of $m_x$, but label value also sees the evidence behind it,
decaying as $(\alpha_{0,x}+1)^{-2}$. With equal total evidence the two rankings
nearly agree. Where it varies they come apart, folded-in labels driving entropy
anti-correlated with value, up to full inversion. Equal evidence is the only
$\alpha_0$ condition giving agreement for every shape, and it fails in any
partially-labelled deployment: label counts alone make $\alpha_0$ heterogeneous.

\section{Experiment 1 auxiliary results}
\label{app:exp1}

All queries in both experiments used the Azure OpenAI deployment
\texttt{gpt-5.6-terra}, API version 2024-12-01-preview.

Escalation numbers on dice. Total predictive entropy tracks judge error at 0.57,
against 0.66 for the epistemic estimate. Escalating on low stated confidence is
indistinguishable from random ($B=10\%$: $+1.8$pp, per-draw 95\% interval $-14$
to $+18$). Ours-versus-entropy regret differences are dominated by single-draw
noise ($+6$pp expected, per-draw $-25$ to $+28$).

\begin{table}[h]
  \caption{Isolation of the estimated components. Spearman correlation between each estimated
    component and each ground truth, with 95\% intervals over label re-draws (real judge) or
    simulated worlds (calibrated). Isolation requires a strong diagonal and near-zero
    off-diagonal. The truth--truth column shows the two ground truths are themselves
    uncorrelated or anti-correlated, so the separation is not inherited from the world.}
  \label{tab:isolation}
  \centering
  \small
  \begin{tabular}{lccccc}
    \toprule
    & \multicolumn{2}{c}{est-aleatoric vs} & \multicolumn{2}{c}{est-epistemic vs} & \\
    \cmidrule(lr){2-3} \cmidrule(lr){4-5}
    world & true disagr. & true error & true disagr. & true error & truth--truth \\
    \midrule
    calibrated simulation & \textbf{0.77} (.76,.78) & 0.13 & $-0.14$
      & \textbf{0.48} (.46,.51) & $-0.03$ \\
    real LLM judge & \textbf{0.45} (.32,.55) & $-0.12$ & 0.08
      & \textbf{0.66} (.51,.70) & $-0.23$ \\
    \bottomrule
  \end{tabular}
\end{table}

\begin{table}[h]
  \caption{The judge's stated confidence is non-monotone in its error. By description tier
    (125 dice each): judge error is the mean Manhattan distance from the judge's predicted
    face distribution to the true one (0--2 scale). Stated confidence is the judge's
    self-reported certainty on its 0--100 scale, and variant spread is the standard deviation of
    the predicted face probabilities across three prompt paraphrases, in probability units
    (0.0002 means the paraphrases agree to the fourth decimal place).}
  \label{tab:confidence}
  \centering
  \small
  \begin{tabular}{lccc}
    \toprule
    tier & judge error & stated confidence & variant spread \\
    \midrule
    none        & 0.53  & \textbf{90.9} & 0.0002 \\
    weak        & 0.54  & \textbf{0.4}  & 0.0008 \\
    directional & 0.41  & 59.1          & 0.0018 \\
    strong      & 0.007 & 14.8          & 0.0006 \\
    \bottomrule
  \end{tabular}
\end{table}

\paragraph{Self-updating versus conjugate updating.}
When shown its own rolls, the judge's self-updated prediction (error 0.258, $n=380$) loses
to conjugate updating of its own prior (adding the observed rolls to its stated
prediction as counts) once the familiarity channel sets the trust level (0.262 with the
channel, 0.307 without): the Bayesian machinery is where the correction lives, not overhead.

\section{Experiment 2 auxiliary results}
\label{app:exp2}

Descriptives. The judge's mean Manhattan error is 0.458 against the 0.158
ceiling, and stated confidence averages 90.3. Item-level Spearman with error is
$-0.13$ for stated confidence and 0.11 for prediction entropy. The fitted trust
is near-uniform, $\mathrm{sd}(\log\hat\lambda) = 0.22$ against the planned
condition's $\ge 0.5$, each prediction worth about five pseudo-labels. The
planned condition tested only the trust side and so predicted no separation, a
mis-specification: the characterisation is about $\alpha_0$, not about
$\lambda$ alone.

Comparisons beyond Table~\ref{tab:esc}. The planned headline, ranking by
$\Delta$, was inconclusive against entropy ($+4.5$ $[-2.2, +9.4]$), and the
counts-only claim was refuted ($+6.9$, grazing zero at $-0.4$). Seven rules are
compared, so no interval stands alone. $I$-ranking beats entropy by $+12.8$
$[+4.3, +20.4]$ at $B=20\%$, with smaller budgets inconclusive. The value
clusters are count-aware 12.1--15.0 and count-blind 6.9--8.3, and $I$ minus
fewest-labels-first is $+0.7$ $[-6.3, +7.1]$.

\begin{table}[h]
  \caption{Escalation on ChaosNLI at $B=10\%$ (311 items). All entries are in Manhattan
    units of error removed, not percentages. Value is the total reduction
    from the purchased labels, summed over the
    escalated set, and advantage is the difference from entropy-ranking on the judge's output,
    paired within replicate. Means over 200 label re-draws, Monte-Carlo standard error in
    parentheses, per-draw 95\% intervals in brackets. Only the $\Delta$ row was the planned
    headline comparison. The remaining advantages are exploratory.}
  \label{tab:escalation}
  \centering
  \small
  \begin{tabular}{lcc}
    \toprule
    escalation rule & value (SE) & advantage over entropy [per-draw 95\%] \\
    \midrule
    ours: mutual information $I$ & 15.0 (0.2) & \textbf{+6.8} $[+1.7, +11.9]$ \\
    fewest-labels-first & 14.3 (0.2) & $+6.1$ $[-0.0, +12.4]$ \\
    ours: label value $\Delta$ \emph{(planned headline)} & 12.7 (0.2)
      & $+4.5$ $[-2.2, +9.4]$ \\
    ours: posterior spread $v$ & 12.1 (0.2) & $+3.9$ $[-1.7, +10.7]$ \\
    random & 8.3 (0.2) & $+0.1$ $[-5.5, +5.6]$ \\
    entropy (judge output) & 8.2 (0.1) & --- \\
    verbalised confidence & 8.1 (0.1) & $-0.1$ $[-5.8, +5.1]$ \\
    entropy (posterior) & 6.9 (0.1) & $-1.2$ $[-6.0, +3.3]$ \\
    \bottomrule
  \end{tabular}
\end{table}

\end{document}